\documentclass[lettersize,journal]{IEEEtran}

\usepackage[caption=false,font=normalsize,labelfont=sf,textfont=sf]{subfig}
\usepackage{stfloats}
\usepackage{url}
\usepackage{graphicx}
\usepackage{cite}
\usepackage{array}
\usepackage{amssymb}
\usepackage{times}
\usepackage{epsfig}
\usepackage{amsmath, amsfonts}
\usepackage{amssymb}
\usepackage{multirow} 
\usepackage{xcolor}
\usepackage{makecell}
\usepackage{textcomp}
\usepackage{longtable}
\usepackage{booktabs}
\usepackage{float}
\usepackage{colortbl}
\usepackage{verbatim}
\usepackage{algorithmic}
\usepackage{algorithm}

\usepackage[capitalize]{cleveref}
\crefname{section}{Sec.}{Secs.}
\Crefname{section}{Section}{Sections}
\Crefname{table}{Table}{Tables}
\crefname{table}{Tab.}{Tabs.}

\begin{document}

\title{L-MAE: Masked Autoencoders are \\Semantic Segmentation Datasets Augmenter}


\author{Jiaru Jia, Mingzhe Liu, Jiake Xie, Xin Chen, Hong Zhang, Feixiang Zhao, Aiqing Yang}

\maketitle
\begin{abstract}
  Generating semantic segmentation datasets has consistently been laborious and time-consuming, particularly in the context of large models or specialized domains(i.e. Medical Imaging or Remote Sensing). Specifically, large models necessitate a substantial volume of data, while datasets in professional domains frequently require the involvement of domain experts. Both scenarios are susceptible to inaccurate data labeling, which can significantly affect the ultimate performance of the trained model. This paper proposes a simple and effective label pixel-level completion method, \textbf{Label Mask AutoEncoder} (L-MAE), which fully uses the existing information in the label to generate the complete label. The proposed model are the first to apply the Mask Auto-Encoder to downstream tasks. In detail, L-MAE adopts the fusion strategy that stacks the label and the corresponding image, namely fuse map. Moreover, since some of the image information is lost when masking the fuse map, direct reconstruction may lead to poor performance. We proposed Image Patch Supplement algorithm to supplement the missing information during the mask-reconstruct process, and empirically found that an average of 4.1\% mIoU can be improved. 
  We conducted a experiment to evaluate the efficacy of L-MAE to complete the dataset. We employed a degraded Pascal VOC dataset and the degraded dataset enhanced by L-MAE to train an identical conventional semantic segmentation model for the initial set of experiments. The results of these experiments demonstrate a performance enhancement of 13.5\% in the model trained with the L-MAE-enhanced dataset compared to the unenhanced dataset.
\end{abstract}


\section{Introduction}
 
 \begin{figure}[h]
 \centering
 \includegraphics[scale=0.076]{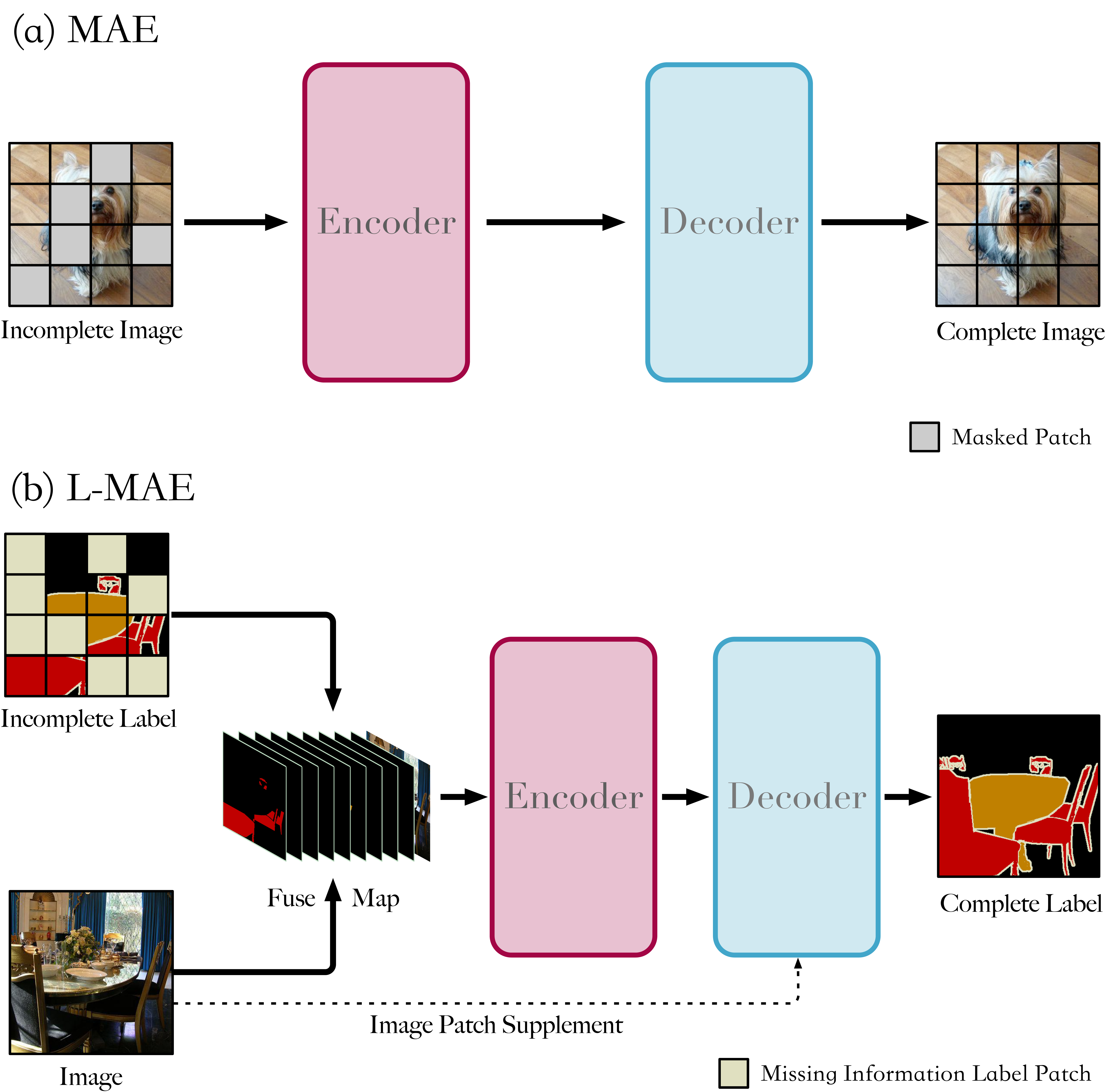}
 \caption{
    \textbf{An illustration between the MAE and our main idea.} (a) MAE conduct a mask and reconstruct strategy on image information, which can learn a feature to complete the incomplete image. (b) To refine the pixel-level on the label of the sementic segmentation dataset, we propose a Label Masked Autoencoder framework. To fully use the input firstly, we present a strategy to fuse the Incomplete Label and the corresponding Image, namely Stack Fusion, to get the Fuse Map. Secondly, we perform the mask-reconstruct pipeline  to get the final accurate prediction. For detail, we design a Image Patch Supplement algorithm to supplement the image information during the pipeline, which can be improved an average of 4.1\% mIoU.
 }
 \label{fig:inference}
 \end{figure}
 
 \begin{figure*}[tp]
   \centering
   \includegraphics[scale=0.115]{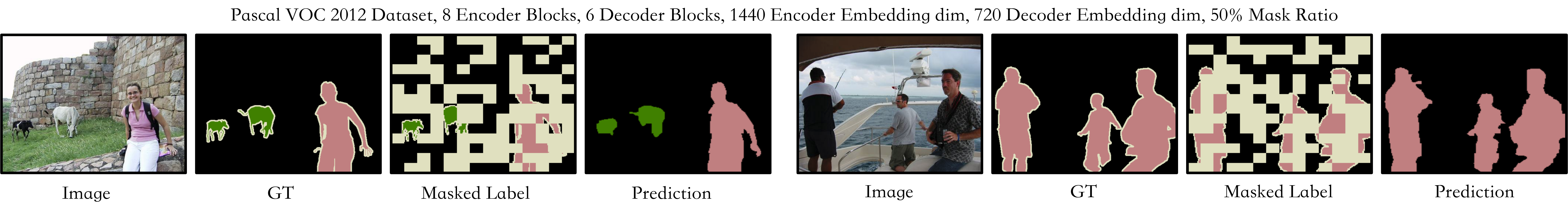}
   \caption{
     \textbf{The performance of the Label Masked Autoencoder.} ``Masked Label" denotes randomly masked complete label, as the ``Prediction" shown, the follow up mask-reconstruct pipeline will complete the masked area.
   }
   \label{fig:output1}
   \end{figure*}
   In the continuous development of semantic segmentation, we have observed that both large-scale deep learning models and research in specific professional fields (such as medical imaging and remote sensing images) have been many research results related to semantic segmentation. These models are tailored for pixel-level semantic analysis of visual data, including images and videos, and offer technical support for diverse applications. To achieve satisfactory performance, large-scale semantic segmentation models rely on extensive datasets, and the models related to professional fields require professionals in the field to participate in creating datasets. However, compared with other tasks, the data labeling work of semantic segmentation models is more complex and challenging, so it is prone to labeling inaccuracies, leading to broken labels.

   To address this issue in generating extensive datasets for professional applications, a common practice involves initially generating a limited number of typical datasets and subsequently employing a trained semi-supervised semantic segmentation model to expand the dataset. Four prevalent categories of methods exist for semi-supervised semantic segmentation models: the first encompasses a GAN-like structure and adversarial training between two networks, one acting as a generator and the other as a discriminator. The second involves consistency regularization methods, these methods include a regularization term in the loss function to minimize the
   differences between different predictions of the same image, which are obtained by applying perturbations
   to the images or to the models involved. The third comprises pseudo-labeling methods, these methods rely on predictions previously made on the unlabeled data with a model trained on the labeled data to obtain pseudo-labels. In this way they are able to include the unlabeled data in the training process. The fourth utilizes contrastive learning techniques, this learning paradigm groups similar elements and separates them from dissimilar elements in a certain representation space, often different from the output space of the models. However, the current approach entails the exclusion of labeled, yet inaccurate, annotations commonly referred to as "broken labels," which may result in inefficient resource utilization. Alternatively, we may incorporate these imprecise and precisely annotated labels in the training dataset for a semi-supervised semantic segmentation model. In that case, this may lead to a decline in the model's performance.

   To fully use the existing broken labels inspired by the Masked Autoencoder model, we proposed the Label Masked Autoencoder. Our model design is divided into a training stage and an inference stage. In the training stage, L-MAE will mask and reconstruct the label. To cover the complex completion scenes in actual situations, we use a mixture of three strategies: Random Mask, Background First Mask, and Label First Mask for the masking strategy. Experiment has shown that the effect of mixed use of the three strategies is significantly better than that of using them alone. At the same time, to allow the model to reconstruct the covered area based on image information, we design the Stack Fuse algorithm to fuse the data of Label and Image. We used label classification based on the layered design idea to highlight the label's information after fusion. Experiments have proven that the fusion strategy used in the model is better than other strategies. Considering the uniformity of the size input to the encoder during the masking step, the model can only mask the entire fused image and label. The circumstance will cover not only the label but also the image. If the model uses 0 to restore the size before inputting it into the decoder, the image information of the masked area will be lost in the decoder. We introduced the Image Patch Supplement (IPS) algorithm in this context. Before transmitting data from the encoder to the decoder, we employed the corresponding image patch to restore the information to its original size. Empirical evidence consistently demonstrates that models incorporating the IPS algorithm outperform those that do not, particularly in terms of completion performance. Lastly, to enable a fair comparison with existing models, we propose using PA-mIoU, a metric that exclusively assesses the mIoU of the region requiring reconstruction. In the context of varying label completion requirements during inference, it is imperative to customize model training by various mask ratios accordingly. To enhance labels, it is noteworthy that regions lacking information predominantly correspond to the background area. After patchifying the label, the proportion of the background area in each label patch is computed. Subsequently, these proportions determine their assignment to the L-MAE model based on the appropriate mask ratio for completion. In the completion phase, the previously utilized hybrid masking strategy from the training stage is abandoned. Instead, patches with a higher background proportion are masked, followed by their subsequent reconstruction.

   To comprehensively assess the performance of L-MAE, we conducted two experiments: the completion performance evaluation and the label enhancement effect evaluation.  In the initial experiment, we compared the mean Intersection over Union (mIoU) achieved by L-MAE in the reconstructed region, denoted as Prediction Area mIoU(PA-mIoU), with that of a fully supervised semantic segmentation model.  This evaluation was carried out on the Pascal VOC and Cityscapes datasets.  Our experimental findings clearly indicate that our mIoU within the reconstructed region surpasses the state-of-the-art (SOTA) results. In the subsequent experiment, we deliberately degraded the labels within the Pascal VOC dataset and utilized these degraded labels to train a conventional semantic segmentation model.  Subsequently, we enriched this dataset using the L-MAE model and trained the same model with the augmented dataset.  The experimental outcomes unambiguously demonstrate that the performance of the model trained with the augmented dataset exhibits a significant improvement when compared to the model trained solely with the degraded dataset.
 
 In summary, our contributions are two-fold:
 
 \begin{itemize}
 \item We propose \textbf{Label Masked Autoencoder}(L-MAE), an Autoencoder-based model, as the first attempt to enhance labels within datasets used for semantic segmentation models. Furthermore, to ensure the model's practical utility, we have developed a multi-mask ratio architecture for the Inference Stage, capable of accommodating varying degrees of completion.
 \item We construct two fundamental algorithms: \textbf{Stack Fuse} and \textbf{Image Patch Supplement}. The Stack Fuse algorithm addresses the challenge of fusing image and label information while ensuring a high-quality completion effect. The Image Patch Supplement algorithm addresses supplementing image information after covering the fuse map.
 \end{itemize}

 The rest of this paper is organized as follows. The second section introduces recent related works. In third section, we describe the proposed method in detail. The fourth section presents the experimental results. The final section concludes the paper.
 
 \section{Related Work}
 \subsection{Vision Transformer and MAE}
 Vision Transformer(ViT) \cite{dosovitskiy2020image}, as the first pure Transformer model for image classification, directly splits an image into nonoverlapping fixed-size patches, adds position embedding, and feeds them to a standard Transformer. It achieves an impressive trade-off between speed and accuracy compared to convolutional neural networks on image classification tasks and solves a problem that has plagued Computer Vision for a long time, how to transmit the image as input into the Transformer and retain the position information of the image. As the model and dataset size grows, there is still no sign of performance saturation. Although ViT works well for image classification, it requires large-scale annotated datasets. Inspired by the application of self-supervised models in NLP, such as BERT \cite{devlin2018bert}, MAE was proposed by He et al. as a self-supervised learning method \cite{alexey2015discriminative, doersch2015unsupervised} for the training of computer vision models. Self-supervised learning mainly uses auxiliary tasks (pretext) to mine its own supervision information from large-scale unsupervised data and trains the network through the constructed supervision information so that it can learn valuable representations for downstream tasks and reduces reliance on large-scale annotated data. MAE proposed an asymmetric encoder-decoder architecture in which the encoder operates only on a subset of visible patches (tokens without masks). Then, there is a lightweight decoder that reconstructs the original image from the latent representation and mask tokens. As a result, the authors found that masking a high proportion of input image patches (e.g. 75\%) turns into a meaningful self-supervised task. Combining these two designs enables efficient training of large models: their models train faster (by 3$\times$ or more) and effectively improve the accuracy.
 \subsection{Sementic Segmentation \& Semi-Supervised Sementic Segmentation Model}

  Semantic segmentation integrates image classification, object detection, and image segmentation, aiming to partition an image into distinct regional blocks, each with a specific semantic meaning, achieved through dedicated techniques.  Subsequently, the semantic category of each regional block is determined, facilitating the progression of semantic reasoning from low-level to high-level information.  Ultimately, the result is a segmented image with pixel-wise semantic annotations.  Presently, the most widely adopted methods for image semantic segmentation rely on convolutional neural networks (CNNs).  Notably, these networks predominantly comprise convolutional layers with two prevalent architectural paradigms: symmetric models (e.g., FCN\cite{long2015fully}, SegNet\cite{badrinarayanan2017segnet}, UNet\cite{ronneberger2015u}) and dilated (e.g., RefineNet\cite{lin2017refinenet}, PSPNet\cite{zhao2017pyramid}, Deeplab series\cite{chen2014semantic,chen2017deeplab,chen2017rethinking}).  Numerous outstanding semantic segmentation models have emerged in the era of the Transformer's prominence.  An exemplar, SegNext\cite{guo2022segnext}, has garnered acclaim for surpassing its predecessors in semantic segmentation performance.  This success can be attributed to its efficient computational design and utilization of the Transformer's encoder structure for feature extraction.

  Semi-supervised semantic segmentation models extract knowledge from labeled data in a supervised way and from unlabeled data in an unsupervised manner, thus reducing the labeling effort required in the fully supervised scenario and achieving better results than in the unsupervised scenario. The commonly used methods include GAN-like structure and adversarial training between the two networks. One as the generator and the other as the discriminator\cite{li2021semantic,mendel2020semi,mittal2019semi}. There are also methods for consistency regularization that include a regularization term in the loss function to minimize the difference between different predictions for the same image\cite{chen2021complexmix,kim2020structured,peng2020deep}. There are also pseudo-labeling methods, which generally rely on predictions previously made on unlabeled data and a model trained on labeled data to obtain pseudo-labels\cite{yang2022st++,zhu2021improving,9889741,9992012,9889681,9785619}. There are also methods based on contrastive learning. This learning paradigm groups and separates similar elements from different elements in a particular representation space\cite{liu2021bootstrapping,alonso2021semi}.

 \subsection{Augmentation Methods}
 Most traditional data enhancement strategies only enhance data. AdvChain \cite{chen2022enhancing} proposes a general GAN-based adversarial data enhancement framework to improve the diversity and effectiveness of data. It generates random photometric and geometric transformations for realistic but challenging imaging variations and extends training data. By optimizing the data augmentation model and segmentation network together during training, examples similar to real-life scenarios are generated to enhance the model's generalization performance in downstream tasks. Their proposed adversarial data augmentation method can be directly used as a module plug-in without relying on the generation network. Sun et al. \cite{sun2021robust} proposed using data enhancement modules. They used two modules: channel random gamma correction and channel random blood vessel enhancement. When given a training color fundus image, the former applies random gamma correction to each channel of the feature map. At the same time, the latter uses a morphological transformation to enhance or reduce only the fine-grained blood vessel parts so that the model can ultimately effectively identify fine-grained blood vessels and better distinguish between global and local interference features. However, these methods cannot fully use the existing image label information when solving images with incomplete annotations, not making the enhancement effect obviously.
 
 \begin{figure*}[!t]
   \centering
   \includegraphics[scale=0.065]{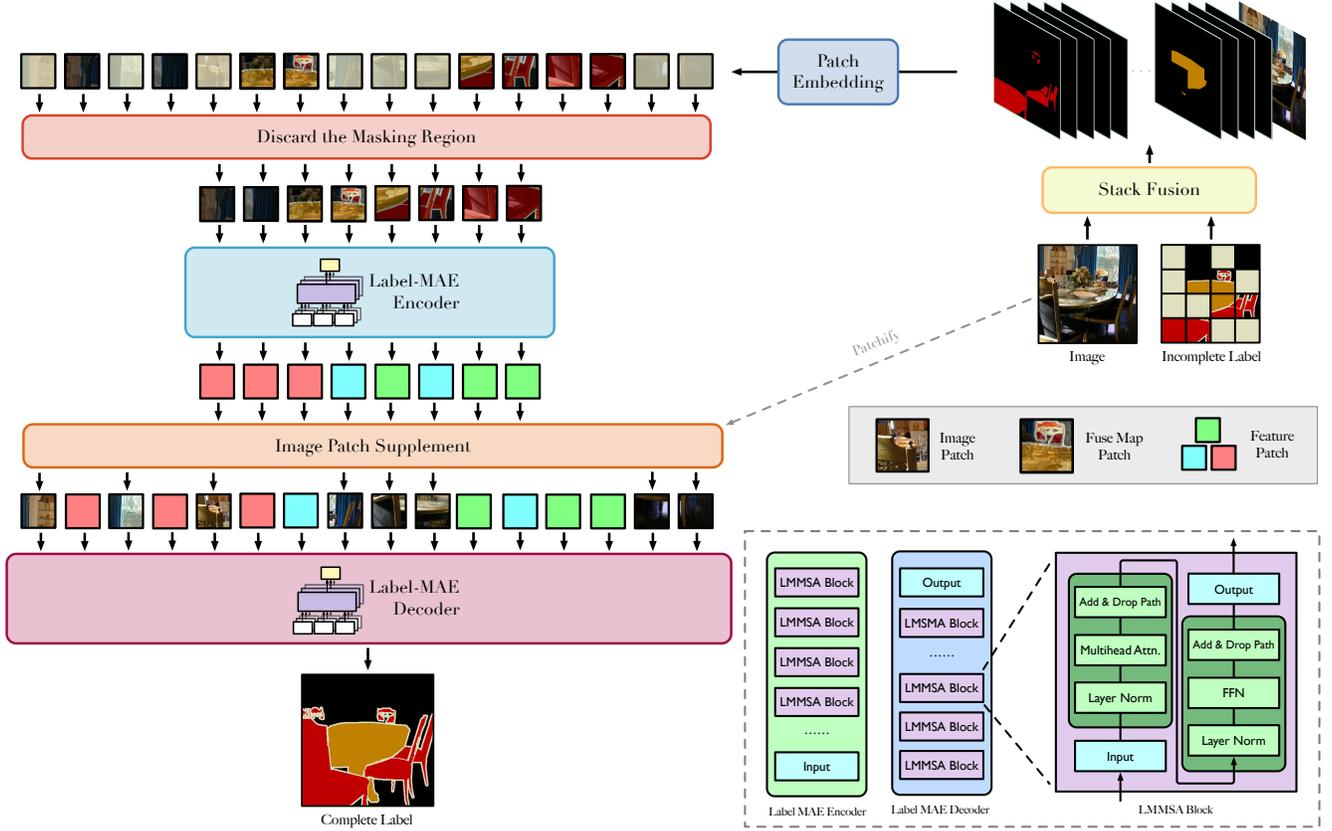}
   \caption{
     \textbf{The overview of the proposed Label Masked Autoencoder(L-MAE) framework.} L-MAE mainly consists of a Stack Fuse module, an L-MAE Encoder, an L-MAE Decoder, and an Image Patch Supplement module. The L-MAE Encoder is applied to the small subset of visible fuse map patches, while the L-MAE Decoder is applied to the masked part. The Image Patch Supplement is used to supplement the image patch to the masked position, the image information here which has lost after performing the Mask Selector module. LMMSA here means Label Masked Autoencoder Multi-Head Attention Module.
   }
   \end{figure*}
   \label{fig:mainfig}
 \section{Label Masked Autoencoder}
 
 The conventional semantic segmentation model and the semi-supervised variant, which enhances the dataset, fall short of addressing potential information gaps within a single label. Our proposed L-MAE model can serve both for completion and semantic segmentation tasks.
 
 In this section, we illustrate the component of our L-MAE. It has three core designs to make it suitable for completing the label, namely Stack Fuse, Image Patch Supplement and the specific strategy for inference phase. The details will be illustrated in the following.
 
 \subsection{Stack Fuse}
 We have tried a variety of methods \cite{bell2016inside, kong2016hypernet} for how to fuse the existing Label and Image: the first is to attach the single-channel Label information directly to the RGB 3-channel Image information in the form of concat. However, in practice, we found that after the subsequent tasks were performed, the results were consistent with the performance of the supervised semantic segmentation model \cite{xie2021segformer, li2022mask, cheng2022masked}, which means the existing Label information was not captured. After the first failure, we thought that the Label information might be submerged in the Image information, and the fusion method was design again. We copied the single-channel Label information three times and inserted it after each RGB channel separately. But the improvement was slight. Finally, we rethought the structure of the semantic segmentation model itself: the network's final output is often a vector of class number dimension, which is not the same as the original Label information dimension. The previous methods may cause the Label information input to the network to be underutilized.  Based on this feature, we finally tried to expand the Label into the class number dimension, concat the image information, and then sent it to the subsequent Encoder-Decoder. We obtained a performance that exceeded the current SOTA (only for the prediction area). 
 
 In detail, for the $Label \in \mathbb{R}^{H\times W\times 1}$ input to the network, we separately extract every class's label, to form a stack of separate Label images, and concatenate them with original Image $I\in \mathbb{R}^{H\times W\times 3}$ as the final Fuse Map $F\in \mathbb{R}^{H\times W\times (N + 3)}$, where the $N$ here is class number.
 The progress can be described as follow, $l_i(i=1,2,...,N)$is the $i$-th class label:
 \begin{gather}
 FuseMap=\text{Concat}(l_1,l_2,...,l_N,image)
 \end{gather}
 
 Before the next step, the Fuse Map will first go through Patch Embedding \cite{gehring2017convolutional}. According to the patch size $p$, the feature $F_f\in \mathbb{R}^{L \times (N+3)}$ can be obtained first, where $L=\frac{H * W}{p * p}$. Then it will be linearly mapped \cite{lusch2018deep} into $F_f\in \mathbb{R}^{L \times \tilde{e}}$ and add position informational to it, where $\tilde{e}$ is encoder embedding dimension, the process can be formulated as follows:
 \begin{gather}
 PE(pos,2i)=F_f + \text{sin}(pos/10000^{2i/\tilde{e}})\ i = 1,2,...,e/2
 \\
 PE(pos,2i+1)=F_f + \text{cos}(pos/10000^{2i/\tilde{e}})\ i = 1,2,...,e/2
 \end{gather}
 where $i$ indexes encoder embedding, and $pos$ indexes each patch.

 \begin{figure*}[!t]
  \centering
  \includegraphics[scale=0.1]{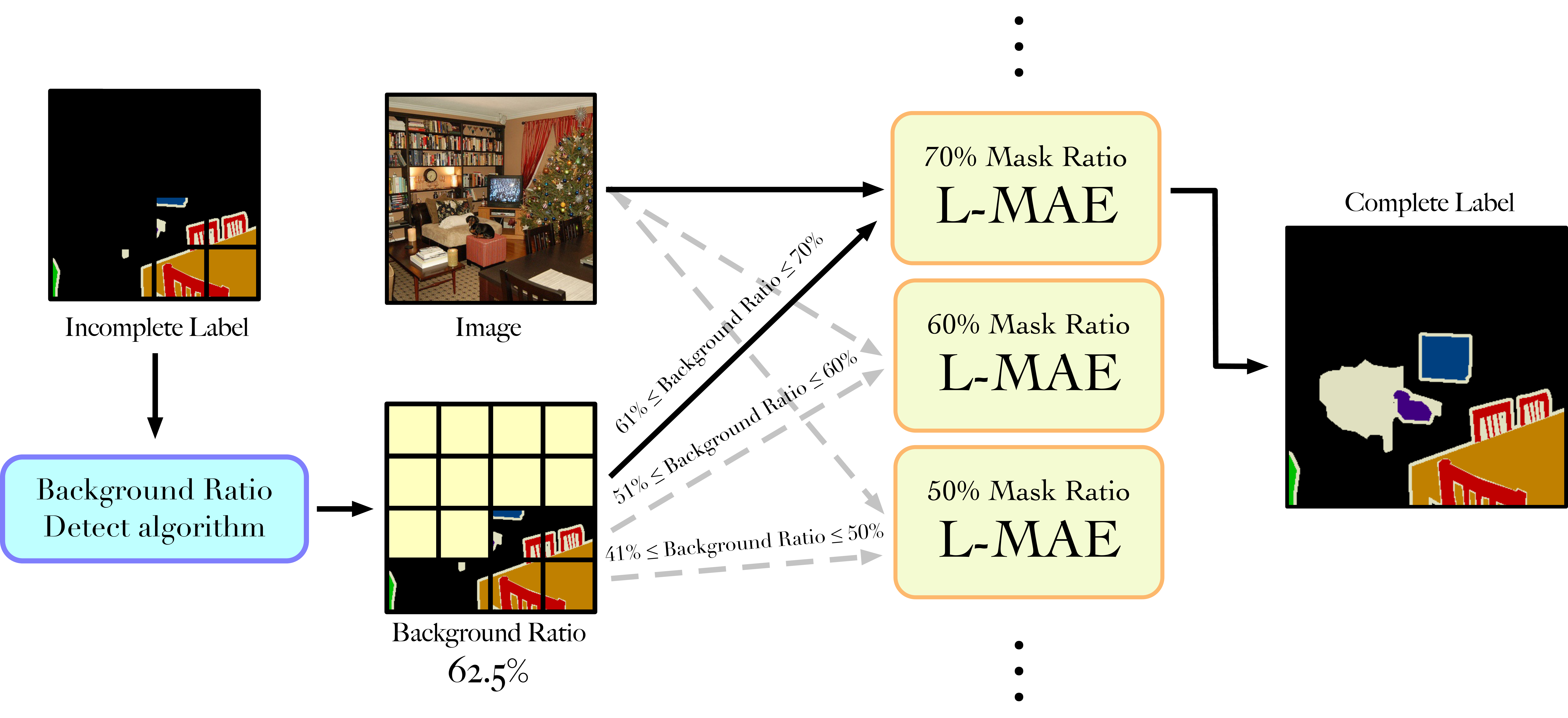}
  \caption{
    \textbf{Inference Phase of Label Masked Autoencoder}. When an image label pair is input, the proportion of the background part of the label will be calculated first, and according to the calculation results, the image label pair will be sent to the L-MAE with the corresponding mask ratio.
  }
  \end{figure*}
  \label{fig:training}

 \subsection{Mask Strategy}
 To enable the model to learn completion methods in various application scenarios, we design three different rules for masking labels during training, namely Random Mask, Background Firstly Mask, and Label Firstly Mask.
 
 The Random strategy will mask the patches in the label indiscriminately. This strategy can cover all possible situations to a great extent. However, when some missing label information is concentrated on a specific object, the random strategy may require more training. Therefore, the Label First strategy will first mask a patch with a high proportion of object tags in a label and then let the model be reconstructed. The strategy enables the model to have the ability to complete when most of the label information of an object is missing. During the training and debugging of the model, We found that when the model completes the label, the original background area is easily predicted to become a nearby object. In order to solve this problem, we design the Background First strategy, which will prioritize discarding areas with a high proportion of background. Under this training strategy, the model's prediction of the background part is strengthened.

 In the actual training process, in order to make the model have the ability to complete an object that is missing most of the annotation information, correctly predict the background part, and complete other general situations except the above, we mixed the Random Mask Strategy, Label Firstly Mask Strategy and Background Firstly Strategy Mask Strategy with a mixing ratio of 1:2:2. Experiments show that when using a mixed strategy, the completion effect of labels is significantly better than the completion effect of using each of the above strategies alone.
 
 \subsection{L-MAE Encoder \& Decoder}
 Our proposed L-MAE model uses the asymmetrical Encoder-Decoder architecture. The encoder is responsible for processing the rest patches, while the decoder process features from all patches.
 
 \textbf{Image Patch Supplement.} L-MAE uses the method of mask and reconstruction to train and learn how to complete an incompletely labeled label. To ensure the uniformity of size, the masking algorithm will be directly based on the fuse map. Since both image and label information are recorded in the fuse map, the direct covering will cover up the Label and Image information simultaneously, which results in a decrease in prediction accuracy due to missing Image information in subsequent completion operations.
 
 So unlike MAE, which uses a normal distribution \cite{khodak2021initialization} for padding, we design an algorithm for it, which named Image Patch Supplement. We first use Patch Embedding that maps each patch into a vector $F_{di}\in \mathbb{R}^{L \times \tilde{d}}$, where $\tilde{d}$ is decoder embedding dimension, and found the patch with the corresponding number on $F_{di}$ according to the discarded patch number recorded before, finally inserted these patches into the $F_e\in \mathbb{R}^{(L-l) \times \tilde{d}}$ output from the Encoder. We name this method Image Patch Supplement. We empirically found that the mIoU of the prediction area is significantly increased after using this method.
 \begin{figure*}[!t]
  \centering
  \includegraphics[scale=0.084]{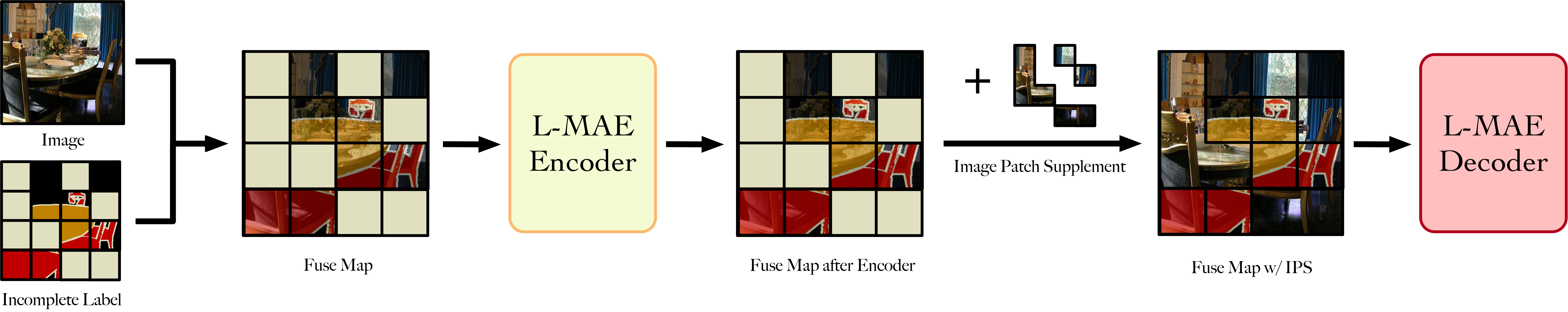}
  \caption{
    The overview of the Image Patch Supplement process: Before the fuse map is sent to the decoder, we use the image corresponding patch information to complete the size of the fuse map instead of using 0. The algorithm can avoid the loss of the image information at the corresponding position due to masking the fuse map by patch.
  }
  \end{figure*}
  \label{fig:mainfig}

 \textbf{Encoder \& Decoder.} For the basic block, we choose to use LMMSA, which means L-MAE Multi-Head Attention module. The Encoder uses N (The default setting is 12) LMMSA Blocks, and the Decoder uses M (The default setting is 8) LMMSA Blocks. For the encoder input or the decoder input $x$, the formula of LMMSA Block is as follows:
 \begin{gather}
 LMMSABlock_1=x+DropPath(MulAttn(x))
 \\
 LMMSABlock_2=x+DropPath(FFN(x))
 \end{gather}
 The Multi-Head Attention (MulAttn) module can perform multiple sets of self-attention processing on the original input sequence, then splicing the results of each set of self-attention to perform a linear transformation to obtain the final output result. Each group of self-attention is a Single-Head Attention. The first step will randomly initialize three vectors, $W^Q, W^K, W^V$. The $Q$, $K$, $V$ can be calculated by:
 \begin{gather}
 Q_i=W_i^QX,\ K_i=W_i^KX,\ V_i=W_i^VX
 \end{gather}
 $Q$, $K$, and $V$ in the above formula are Query, Key, and Value, respectively. The input data $x$ will use $Q$ to retrieve the similar parts of the $K$ by multiplication, the correlation coefficient will be calculated through softmax, and finally, the $V$ will be calculated. By weighted summation, an output vector can be obtained, and the formula is as follows:
 \begin{gather}
 Z_i=softmax(\frac{Q_iK_i^T}{\sqrt{d_k}})V_i
 \\
 MulAttn(Q,K,V)=concat(Z_1,...,Z_h)W^O
 \end{gather}
 where $h$ represents the number of heads, $W^O \in \mathbb{R}^{hd_v \times d_{model}}$, ${W^Q \in \mathbb{R}^{d_{model} \times d_k}}$ , ${W^K \in \mathbb{R}^{d_{model} \times d_k}}$ and ${W^V_i \in \mathbb{R}^{d_{model} \times d_v}}$ are projected parameter matrices. $Z_i$ is the output vector of each attention head. In original ViT, they employ $h=8$ parallel attention layers and for each heads they use $d_k = d_v = d_{model}/h = 64$. Although there are many heads in the multi-head attention mechanism, the final complexity is similar to that of single-head attention.
 
 A Feed-Forward Neural networks(FFN) are applied to the encoder and decoder in the last part, followed by a layer normalization \cite{ba2016layer}. FFN are calculated as follows:
 \begin{gather}
 FFN(x)=ReLU(W_1x+b_1)W_2+b_2
 \end{gather}
 Although the same linear transformation is used for different positions, different parameters are used in feed-forward networks from one layer to another.

 \subsection{Inference Phase}
 In inference phase, we train various Mask Ratio L-MAE models. When an Image-Label pair is input, the Label in it will be detected, and based on the proportion of its background part, it will be input to Enhancement is performed in the L-MAE of the Mask Ratio that conforms to this ratio. During the enhancement process, the Background First algorithm will be used to mask the block parts that are Background First. After passing through the Image Patch Supplement module and Decoder, the initially missing parts can be completed.

 \subsection{Loss Design}
 For the pipeline generated label $x$ and the given original label $y$, a cross entropy loss \cite{polat2022class} is adopted to optimize the relationship between two modalities, where $x$ is encouraged to be similar with its corresponding $y$ and dissimilar with other irrelatvant $y$. For detail, to solve the problem of performance degradation caused by the different proportions of each class in the segmentation task, we design a weight calculation algorithm, so that each class can calculate the loss more balanced according to the proportion in the dataset. For the input $x$ and $y$, the whole loss calculate process can be formulated as:
 \begin{gather}
 w_i=1 - 1/(1+\beta \cdot exp(-\frac{x_i-E(x)}{V(x)+\xi})^\gamma)
 \\
 l_n=-w_{i}\cdot log\frac{exp(x_{n,y_n})}{\sum_{i=1}^{C}exp(x_{n,i})} \cdot 1\{y_n  \neq \hat{y}_n\}
 \\
 \ell(x,y)=\sum_{i=1}^{\mathcal{N}+1}l_i \cdot \frac{1}{\sum_{n=1}^{\mathcal{N}+1} w_{y_{n}} \cdot 1\{y_{n} \neq \hat{y}_n\}}
 \end{gather}
 Where $i\in \mathcal{N}$, $\hat{y}_n$ is the ignore index, $\ell$ is the loss function.
 For detail, $\beta$ and $\gamma$ are the learnable bias, $\mathcal{N}$ means the number of class, which the $\mathcal{N}+1$ is the output vector dimension, $x_n$ and $y_n$ here is the dimension $n$ of the prediction $x$ and the real label $y$ , $w_{y_n}$ means for the demension n, the loss weight of class $y_n$, ignore index $\hat{y}_n$ is the background class, which not take account in the loss calculation.

 \section{Experiment}
 In this section, we introduce the technical details of applying our proposed L-MAE first. Then we design and conduct comparative experiments on Pascal VOC 2012 and Cityscapes datasets to compare the performance differences between L-MAE and (semi-)supervised semantic segmentation models. Finally, we design a set of extensive ablation experiments to analyze the effect of Encoder Block, Decoder Block, Encoder Embedding Dimension, Decoder Embedding Dimension, and Mask Ratio on performance in L-MAE.
 \subsection{Experiment Setup}
 Our L-MAE is built using the Pytorch framework and the Adam optimizer \cite{kingma2014adam} with 0.9 of momentum, and weight decay is set to 0.0001. Further, the Reduce Learning Rate On Plateau strategy with min mode is used, which reduces the learning rate with a specified factor after a preset number of patient rounds, which is decided by threshold of loss dropping. In actual use, we set these parameters as follows: 5 of the patient rounds, 0.001 of the threshold, and 0.8 of the factor.
 The experiments were performed using Nvidia Tesla A40 for 400 epochs on Pascal VOC and Cityscapes. On Pascal VOC, the Image and Label are randomly cropped to $448 \times 448$, and then scaled to $224 \times 224$, while in Cityscapes which is randomly cropped to $448 \times 448$. During training, a batch size of 24 was used in Pascal VOC, and a batch size of 48 was used in Cityscapes.
 
 \subsection{PA-mIoU} 
 It is unfair to use the general global area mIoU solely to other semantic segmentation models in the comparative experiment, which may lead to low efficient of measuring the network performance. So we design a new evaluation metric: \textbf{Predict Area mean Intersection overUnion} (PA-mIoU), which will only focus on the mIoU in discarded area. In the specific implementation, the number list $i \in \mathbb{R}^{l}$ of the patch is discarded according to the Mask Selector, and a mask $m \in \mathbb{R}^{H \times W}$ that can cover the entire label is produced. Then according to whether each position in the mask stores 0 or 1 (0 is the reserved area, 1 is the discarded area), to determine whether the result of the label in the corresponding position is counted.
 
 \begin{table*}[t]
  \renewcommand{\arraystretch}{1.3}
  \caption{Comparison with the supervised sementic segmentation state-of-the-art approach on Pascal VOC 2012  and Cityscapes dataset. For fairly, \textbf{our results use the PA-mIoU metric}. ``m" denotes the Mask Ratio, while our L-MAE choose the hyper-parameter combination of Encoder Block is 8, Decoder Block is 6, Encoder Embedding Dimension is 1440, and Decoder Embedding Dimension is 720.\\}
  \resizebox{\textwidth}{!}
  {
  \begin{tabular}{m{2.7cm}<{\raggedright}|m{0.6cm}<{\centering}m{0.6cm}<{\centering}m{0.6cm}<{\centering}m{0.6cm}<{\centering}m{0.7cm}<{\centering}m{0.6cm}<{\centering}m{0.6cm}<{\centering}m{0.6cm}<{\centering}m{0.6cm}<{\centering}m{0.6cm}<{\centering}m{0.6cm}<{\centering}m{0.6cm}<{\centering}m{0.6cm}<{\centering}m{0.7cm}<{\centering}m{0.8cm}<{\centering}m{0.6cm}<{\centering}m{0.6cm}<{\centering}m{0.5cm}<{\centering}m{0.6cm}<{\centering}m{0.5cm}<{\centering}|m{0.8cm}<{\centering}}
  
  \Xhline{1.4pt}
    
  \textbf{Methods} & \textbf{aero} & \textbf{bike} & \textbf{bird} & \textbf{boat} & \textbf{bottle} & \textbf{bus} & \textbf{car} & \textbf{cat} & \textbf{chair} & \textbf{cow} & \textbf{table} & \textbf{dog} & \textbf{horse} & \textbf{mbike} & \textbf{person} & \textbf{plant} & \textbf{sheep} & \textbf{sofa} & \textbf{train} & \textbf{tv} & \textbf{mIoU}\\ 
  
  \Xhline{0.8pt}
  
  RefineNet\cite{lin2017refinenet} & 95.0 & 73.2 & 93.5 & 78.1 & 84.8 & 95.6 & 89.8 & 94.1 & 43.7 & 92.0 & 77.2 & 90.8 & 93.4 & 88.6 & 88.1 & 70.1 & 92.9 & 64.3 & 87.7 & 78.8 & 84.2\\
  
  ResNet38\cite{he2016deep} & 96.2 & 75.2 & 95.4 & 74.4 & 81.7 & 93.7 & 89.9 & 92.5 & 48.2 & 92.0 & 79.9 & 90.1 & 95.5 & 91.8 & 91.2 & 73.0 & 90.5 & 65.4 & 88.7 & 80.6 & 84.9\\
  
  PSPNet\cite{zhao2017pyramid}  & 95.8 & 72.7 & 95.0 & 78.9 & 84.4 & 94.7 & 92.0 & 95.7 & 43.1 & 91.0 & 80.3 & 91.3 & 96.3 & 92.3 & 90.1 & 71.5 & 94.4 & 66.9 & 88.8 & 82.0 & 85.4\\
  
  Deeplabv3\cite{chen2017rethinking} & 96.4 & 76.6 & 92.7 & 77.8 & 87.6 & 96.7 & 90.2 & 95.4 & 47.5 & 93.4 & 76.3 & 91.4 & 97.2 & 91.0 & 92.1 & 71.3 & 90.9 & 68.9 & 90.8 & 79.3 & 85.7\\
  
  EncNet\cite{zhang2018context} & 95.3 & 76.9 & 94.2 & 80.2 & 85.3 & 96.5 & 90.8 & 96.3 & 47.9 & 93.9 & 80.0 & 92.4 & 96.6 & 90.5 & 91.5 & 70.9 & 93.6 & 66.5 & 87.7 & 80.8 & 85.9\\
  
  DFN\cite{yu2018learning} & 96.4 & 78.6 & 95.5 & 79.1 & 86.4 & 97.1 & 91.4 & 95.0 & 47.7 & 92.9 & 77.2 & 91.0 & 96.7 & 92.2 & 91.7 & 76.5 & 93.1 & 64.4 & 88.3 & 81.2 & 86.2\\
  
  SDN\cite{fu2019stacked} & 96.9 & 78.6 & 96.0 & 79.6 & 84.1 & 97.1 & 91.9 & 96.6 & 48.5 & 94.3 & 78.9 & 93.6 & 95.5 & 92.1 & 91.1 & 75.0 & 93.8 & 64.8 & 89.0 & 84.6 & 86.6\\
  
  Deeplabv3+\cite{chen2018encoder} & 97.0 & 77.1 & 97.1 & 79.3 & 89.3 & 97.4 & 93.2 & 96.6 & 56.9 & 95.0 & 79.2 & 93.1 & 97.0 & 94.0 & 92.8 & 71.3 & 92.9 & 72.4 & 91.0 & 84.9 & 87.8\\
  
  ExFuse\cite{zhang2018exfuse} & 96.8 & \cellcolor{yellow}\textbf{80.3} & 97.0 & 82.5 & 87.8 & 96.3 & 92.6 & 96.4 & 53.3 & 94.3 & 78.4 & 94.1 & 94.9 & 91.6 & 92.3 & 81.7 & 94.8 & 70.3 & 90.1 & 83.8 & 87.9\\
  
  MSCI\cite{lin2018multi} & 96.8 & 76.8 & 97.0 & 80.6 & 89.3 & 97.4 & 93.8 & 97.1 & 56.7 & 94.3 & 78.3 & 93.5 & 97.1 & 94.0 & 92.8 & 72.3 & 92.6 & 73.6 & 90.8 & 85.4 & 88.0\\
  
  MRFM\cite{yuan2020multi} & \cellcolor{yellow}\textbf{97.1} & 78.6 & \cellcolor{yellow}\textbf{97.1} & 80.6 & 89.7 & \cellcolor{yellow}\textbf{97.3} & \cellcolor{yellow}\textbf{93.6} & \cellcolor{yellow}\textbf{96.7} & 59.0 & \cellcolor{yellow}\textbf{95.4} & 81.1 & 93.2 & \cellcolor{yellow}\textbf{97.5} & \cellcolor{yellow}\textbf{94.2} & \cellcolor{yellow}\textbf{92.9} & 72.3 & 93.1 & 74.2 & 91.0 & 85.0 & 88.4\\
  
  \Xhline{0.8pt}
  
  \textbf{L-MAE w/ m=0.6} & 89.5 & 58.8 & 92.3 & 86.4 & 91.3 & 94.9 & 89.6 & 95.3 & 79.1 & 93.7 & 89.6 & 93.4 & 91.0 & 90.5 & 89.4 & 85.6 & 96.1 & 93.0 & 96.0 & 87.5 & \cellcolor{cyan}\textbf{89.1}\\
  
  \textbf{L-MAE w/ m=0.5} & 89.9 & 64.5 & 91.4 & \cellcolor{yellow}\textbf{89.2} & \cellcolor{yellow}\textbf{92.1} & 95.9 & 90.6 & 96.4 & \cellcolor{yellow}\textbf{82.1} & 94.6 & \cellcolor{yellow}\textbf{91.2} & \cellcolor{yellow}\textbf{94.7} & 94.3 & 92.4 & 91.1 & \cellcolor{yellow}\textbf{87.8} & \cellcolor{yellow}\textbf{97.8} & \cellcolor{yellow}\textbf{94.0} & \cellcolor{yellow}\textbf{96.5} & \cellcolor{yellow}\textbf{93.4} & \cellcolor{green}\textbf{91.0}\\
  
  \Xhline{1.4pt}
        
  \end{tabular}
  }
  \end{table*}

  \begin{table*}[t]
     \renewcommand{\arraystretch}{1.3}
    \resizebox{\textwidth}{!}
    {
      \begin{tabular}{m{2.7cm}<{\raggedright}|m{0.7cm}<{\centering}m{0.7cm}<{\centering}m{0.7cm}<{\centering}m{0.7cm}<{\centering}m{0.7cm}<{\centering}m{0.7cm}<{\centering}m{0.7cm}<{\centering}m{0.7cm}<{\centering}m{0.7cm}<{\centering}m{0.7cm}<{\centering}m{0.7cm}<{\centering}m{0.7cm}<{\centering}m{0.7cm}<{\centering}m{0.5cm}<{\centering}m{0.6cm}<{\centering}m{0.5cm}<{\centering}m{0.6cm}<{\centering}m{0.8cm}<{\centering}m{1.0cm}<{\centering}|m{0.8cm}<{\centering}}
      
      \Xhline{1.4pt}
        
      \textbf{Methods} & \textbf{road} & \textbf{swalk} & \textbf{build.} & \textbf{wall} & \textbf{fence} & \textbf{pole} & \textbf{tlight} & \textbf{tsign} & \textbf{veg.} & \textbf{terrain} & \textbf{sky} & \textbf{person} & \textbf{rider} & \textbf{car} & \textbf{truck} & \textbf{bus} & \textbf{train} & \textbf{mcycle} & \textbf{bicycle} & \textbf{mIoU}\\ 
      
      \Xhline{0.8pt}
      
      VPLR\cite{zhu2019improving} & 98.8 & 87.8 & 94.2 & 64.1 & 65.0 & 72.4 & 79.0 & 82.8 & 94.2 & 74.0 & 96.1 & 88.2 & 75.4 & 96.5 & 78.8 & 94.0 & 91.6 & 73.7 & 79.0 & 83.5\\
      
      HRNet-OCR\cite{yuan2020object}  & 98.8 & 88.3 & 94.1 & 66.9 & 66.7 & 73.3 & 80.2 & 83.0 & 94.2 & 74.1 & 96.0 & 88.5 & 75.8 & 96.5 & 78.5 & 91.8 & 90.1 & 73.4 & 79.3 & 83.7\\
      
      P-Deeplab\cite{cheng2020panoptic} & 98.8 & 88.1 & 94.5 & 68.1 & 68.1 & 74.5 & 80.5 & 83.5 & 94.2 & 74.4 & 96.1 & 89.2 & 77.1 & 96.5 & 78.9 & 91.8 & 89.1 & 76.4 & 79.3 & 84.2\\
      
      iFLYTEK-CV & 98.8 & 88.4 & 94.4 & 68.9 & 66.8 & 73.0 & 79.7 & 83.3 & 94.3 & 74.3 & 96.0 & 88.8 & 76.3 & 96.6 & 84.0 & \cellcolor{yellow}\textbf{94.3} & 91.7 & 74.7 & 79.3 & 84.4\\
      
      SegFix\cite{yuan2020segfix} & 98.8 & 88.3 & 94.3 & 67.9 & 67.8 & 73.5 & 80.6 & 83.9 & 94.3 & 74.4 & 96.0 & 89.2 & 75.8 & \cellcolor{yellow}\textbf{96.8} & 83.6 & 94.1 & 91.2 & 74.0 & 80.0 & 84.5\\
      
      HMSA\cite{tao2020hierarchical} & \cellcolor{yellow}\textbf{99.0} & 89.2 & 94.9 & 71.6 & 69.1 & 75.8 & \cellcolor{yellow}\textbf{82.0} & \cellcolor{yellow}\textbf{85.2} & 94.5 & 75.0 & \cellcolor{yellow}\textbf{96.3} & \cellcolor{yellow}\textbf{90.0} & 79.4 & 96.9 & 79.8 & 94.0 & 85.8 & 77.4 & 81.4 & 85.1\\
      
      \Xhline{0.8pt}
      
      \textbf{L-MAE w/ m=0.6} & 97.4 & 89.1 & 94.1 & 89.2 & 90.7 & 74.8 & 66.0 & 71.7 & 93.9 & 88.6 & 94.5 & 81.7 & 73.8 & 93.6 & 89.6 & 89.8 & 90.3 & 79.3 & 78.0 & \cellcolor{cyan}\textbf{85.6}\\
  
      \textbf{L-MAE w/ m=0.5} & 98.0 & \cellcolor{yellow}\textbf{91.9} & \cellcolor{yellow}\textbf{95.8} & \cellcolor{yellow}\textbf{91.9} & \cellcolor{yellow}\textbf{93.0} & \cellcolor{yellow}\textbf{81.4} & 74.4 & 78.6 & \cellcolor{yellow}\textbf{95.7} & \cellcolor{yellow}\textbf{91.1} & 95.8 & 86.2 & \cellcolor{yellow}\textbf{79.7} & 95.3 & \cellcolor{yellow}\textbf{91.9} & 91.7 & \cellcolor{yellow}\textbf{92.1} & \cellcolor{yellow}\textbf{83.6} & \cellcolor{yellow}\textbf{82.9} & \cellcolor{green}\textbf{86.4}\\
      
      \Xhline{1.4pt}
            
      \end{tabular}
    }
  \end{table*}

 \begin{table}
   \centering
   \renewcommand{\arraystretch}{1.3}
   \caption{Comparison with semi-supervised semantic segmentation state-of-the-art U2PL on Pascal VOC 2012. \textbf{The metric mIoU  for L-MAE is PA-mIoU for fairly.}\\}
   \scalebox{0.9}
   {
   \begin{tabular}{c|cc}
       \Xhline{1.4pt}
       \textbf{Method} & \textbf{Mask\ Ratio} & \textbf{mIoU} \\
       \Xhline{0.8pt} 
       \multirow{3}{*}{U2PL} & 87.5\% & 79.01 \\
       \cline { 2 - 3 } & 75\% & 79.30 \\
       \cline { 2 - 3 } & 50\% & 80.5 \\
       \Xhline{0.8pt} 
       \multirow{3}{*}{S4MC} & 87.5\% & 79.67 \\
       \cline { 2 - 3 } & 75\% & 79.85 \\
       \cline { 2 - 3 } & 50\% & 81.1 \\
       \Xhline{0.8pt} 
       \multirow{3}{*}{L-MAE(w/ IPS)} & 80.0\% & 85.5 \\
       \cline { 2 - 3 } & 70\% & 90.1 \\
       \cline { 2 - 3 } & 50\% & 91.3 \\
       \Xhline{1.4pt}
   \end{tabular}
   }
 \end{table}

 \begin{table}
  \centering
  \renewcommand{\arraystretch}{1.9}
  \caption{Comparison with serveral different Stack Fuse Method, the Directly Concat here is concat the label to the back of the Image RGB 3 layers directly, the Insert Concat here is copy the label 3 times, and insert them between the RGB 3 layers, the Layer Concat is to layer labels by category.\\}
  \scalebox{0.9}
  {
  \begin{tabular}{c|cc}
      \Xhline{1.4pt}
      \textbf{Method} & \textbf{PA-mIoU} & \textbf{mIoU} \\
      \Xhline{0.8pt} 
      Directly Concat & 72.4\% & 74.8\% \\
      \Xhline{0.8pt} 
      Insert Concat & 75.6\% & 77.9\% \\
      \Xhline{0.8pt} 
      Directly Concat & 94.6\% & 91.3\% \\
      \Xhline{1.4pt}
  \end{tabular}
  }
\end{table}

\subsection{Label Augmentation Experiment}
We conducted the Label Augmentation Experiment to assess whether enhancing the dataset with L-MAE yields performance improvements in conventional semantic segmentation models.  Within this experiment, we intentionally degraded the Pascal VOC dataset by randomly obscuring 50\% of the data.  We then employed this degraded dataset to train FCN and UNet models for 300 iterations.  Subsequently, we applied L-MAE to enhance the degraded dataset and employed this improved dataset to retrain the FCN and UNet models for another 300 iterations.  The results demonstrate notable enhancements in the performance of the trained FCN and UNet models on the test set, with improvements of 13.4\% and 11.7\%, respectively, compared to the original dataset.  These findings strongly affirm the effectiveness of L-MAE in practical scenarios.

 \subsection{Comparative Experiments}
 
 In comparative experiments, we use several excellent supervised semantic segmentation models to compare with our Label Masked Autoencoder. By comparing the PA-mIoU of L-MAE with the mIoU of other models, we found that as the Mask Ratio decreases, the PA-mIoU will continue to rise, and exceeds the existing state-of-the-art model when the Mask Ratio is 50\%.
 \subsubsection{Results on Pascal VOC 2012}
 Pascal VOC 2012 is an upgraded version of the Pascal VOC 2007 dataset, with a total number of 11530 images. For the segmentation task, the train/val of VOC2012 contains all images from 2007 to 2012, with 2913 images totaling, 2513 of which used as the training set and 400 images are used as the validation set. For the comparison with other general sementic segmentation models, when the mask ratio is set to 50\%, our L-MAE can achieve \textbf{94.6\%} Global mIoU, and  \textbf{91.0\%} PA-mIoU; when the mask ratio is set to 60\%, our L-MAE can achieve \textbf{92.6\%} mIoU, and \textbf{89.1\%} PA-mIoU. As shown in the top of the Table 1, the proposed method outperforms the current conventional performance. As the comparison with the semi-supervised sementic segmentation model, we chose U2PL and S4MC as comparison models. According to the table, the performance of mIoU under a similar Mask Ratio is 5\% or more different from L-MAE. Whether from the comparison data of the supervised model or the comparison data with the semi-supervised model, the performance superiority of the L-MAE model can be proved.
 
 \subsubsection{Results on Cityscapes} 
 Cityscapes dataset is collected from 50 cities in Germany and nearby countries, including street scenes in three seasons of spring, summer and autumn, with 5000 images of 27 cities with pixel-level semantic and instance annotation. For comparing, we set the Mask Ratio to 50\% and 60\%, respectively. As the result, our L-MAE can achieve \textbf{90.5\%} Global mIoU, and \textbf{86.4\%} PA-mIoU, when the Mask Ratio is 50\%, while the \textbf{89.0\%} Global mIoU, and \textbf{85.6\%} PA-mIoU in 60\% Mask Ratio. As shown in the bottom of the Table 1, L-MAE surpasses the legacy SOTA. For comparison with the semi-supervised semantic segmentation model, we chose the current SOTA model U2PL. As the Table 2 shows, whether it is under similar Mask Ratio settings (87.5\% with 80\% and 75\% with 70\%) or the same setting (50\%), there are significant advantages in mIoU indicators.

 \begin{table}
   \renewcommand{\arraystretch}{1.2}
   \caption{Comparison between different hyper-parameter combinations. ``$EB$" is Encoder Block, ``$DB$" is Decoder Block, ``$ED$" is Encoder Embedding Dimension, ``$DD$" is Decoder Embedding Dimension. We also calculate the parameters(M) and the FLOPs(G) under various settings. Moreover, the calculation of the mIoU and the PA-mIoU include the background class.\\}
   \resizebox{\linewidth}{!}
   {
   \begin{tabular}{c|c|c|c|c|c|c|c}
       \Xhline{1.4pt}
       \textbf{$EB$} & \textbf{$DB$} & \textbf{$ED$} & \textbf{$DD$} & \textbf{\#params.(M)} & \textbf{FLOPs(G)} & \textbf{mloU} & \textbf{PA-mloU} \\
       \Xhline{0.8pt} \multirow{2}{*}{12} & \multirow{2}{*}{8} & 1024 & 512 & $186 \mathrm{M}$ & $21 \mathrm{G}$ & $94.1$ & $90.5$ \\
       \cline { 3 - 8 } & & 1440 & 720 & $362 \mathrm{M}$ & $42 \mathrm{G}$ & $94.1$ &$90.4\downarrow$ \\
       \Xhline{0.8pt} \multirow{2}{*}{8} & \multirow{2}{*}{6} & 1024 & 512 & $129 \mathrm{M}$ & $15 \mathrm{G}$ & $94.1$ & $90.3$ \\
       \cline { 3 - 8 } & & 1440 & 720 & $250 \mathrm{M}$ & $29 \mathrm{G}$ & $94.6\uparrow$ & $91.3\uparrow$ \\
       \Xhline{0.8pt} \multirow{2}{*}{6} & \multirow{2}{*}{4} & 1024 & 512 & $98 \mathrm{M}$ & $11 \mathrm{G}$ & $94.2$ & $90.4$ \\
       \cline { 3 - 8 } & & 1440 & 720 & $188 \mathrm{M}$ & $22 \mathrm{G}$ & $94.6\uparrow$ & $91.2\uparrow$ \\
       \Xhline{1.4pt}
   \end{tabular}
   }
 \end{table}

 \begin{table}
  \centering
  \renewcommand{\arraystretch}{1.9}
  \caption{Compare the performance difference between ordinary semantic segmentation networks trained using the L-MAE enhancement method and unenhanced ones.\\}
  \scalebox{0.9}
  {
  \begin{tabular}{c|c|c}
      \Xhline{1.4pt}
      \textbf{Network} & \textbf{mIoU w/ L-MAE} & \textbf{mIoU w/o L-MAE} \\
      \Xhline{0.8pt} 
      FCN & 43.5\% & 57.9\% \\
      \Xhline{0.8pt} 
      UNet & 59.5\% & 71.2\% \\
      \Xhline{1.4pt}
  \end{tabular}
  }
\end{table}

 \subsection{Ablation Study}
 To evaluate the impact and performance of each component in our model, we evaluate their effectiveness in this section. The Pascal VOC dataset(224 input size) is chosen to evaluate the model.
 
 \subsubsection{Parameter Setting Analysis}
 
 As shown in Table 3, we explore the performance of the L-MAE under different parameters. There are 4 main hyper-parameters for the network: encoder block number($EB$), decoder block number($DB$), encoder embedding dimension($ED$), decoder embedding dimension($DD$). We can find that when the block number is consistent, the performence will get worse as the embedding dimension fall. Nevertheless, expect for the case $EB$=12, $DB$=8, when using $ED$=1440 and $DD$=720, its PA-mIoU is lower than when $ED$=1024 and $DD$=512. However, the global mIoU keeps the same, which means that the mIoU of the reserved area has risen. Based on the table results, we choose the combination of $EB$=8, $DB$=6, $ED$=1440, and $DD$=720.
 
 \begin{figure}[tp]
   \centering
   \caption{Ablation study on various Mask Ratio, and whether or not to use the Image Patch Supplement(IPS) algorithm. The calculation of the mIoU and the PA-mIoU include the background class. "m" means Mask Ratio.\\}
   \includegraphics[scale=0.25]{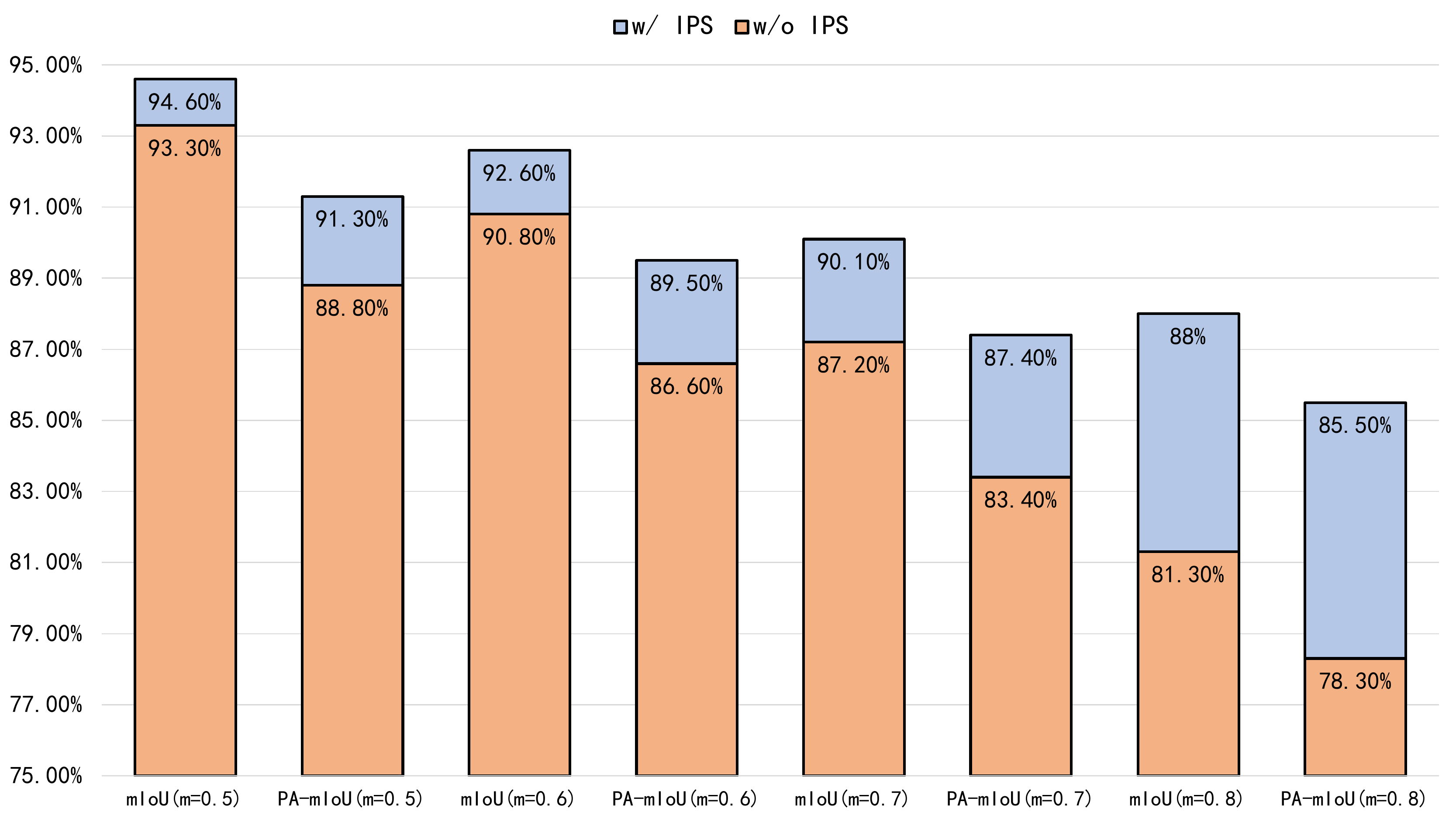}
   \label{fig:ips}
   \end{figure}
   \begin{figure}[!tp]
    \centering
    \caption{
      Label Augmentation experiment study pipeline, the plain segmentic segmentation network will calculate the loss with the L-MAE regenerated label.\\
    }
    \includegraphics[scale=0.07]{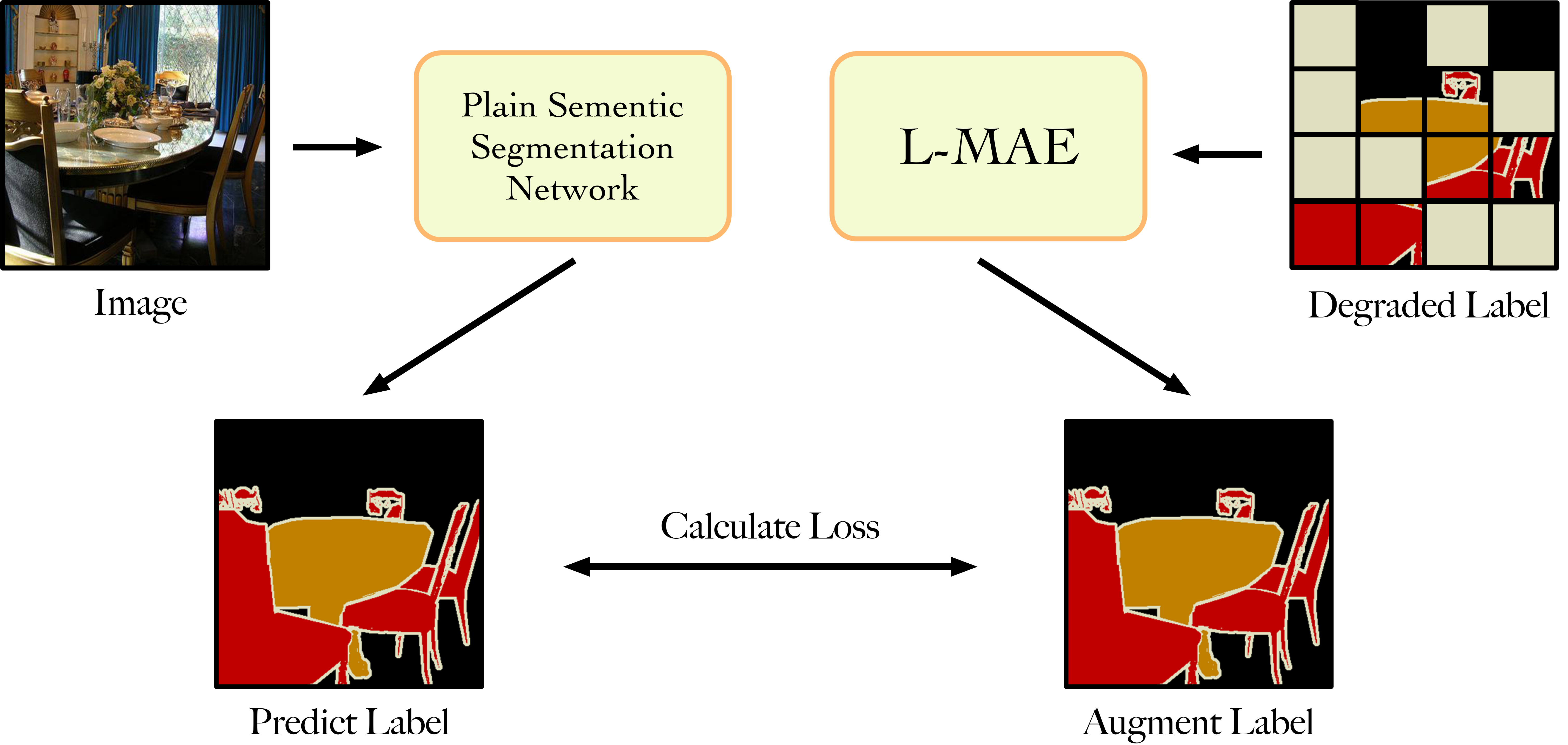}
    \end{figure}
  \begin{figure*}[tp]
    \centering
    \includegraphics[scale=0.097]{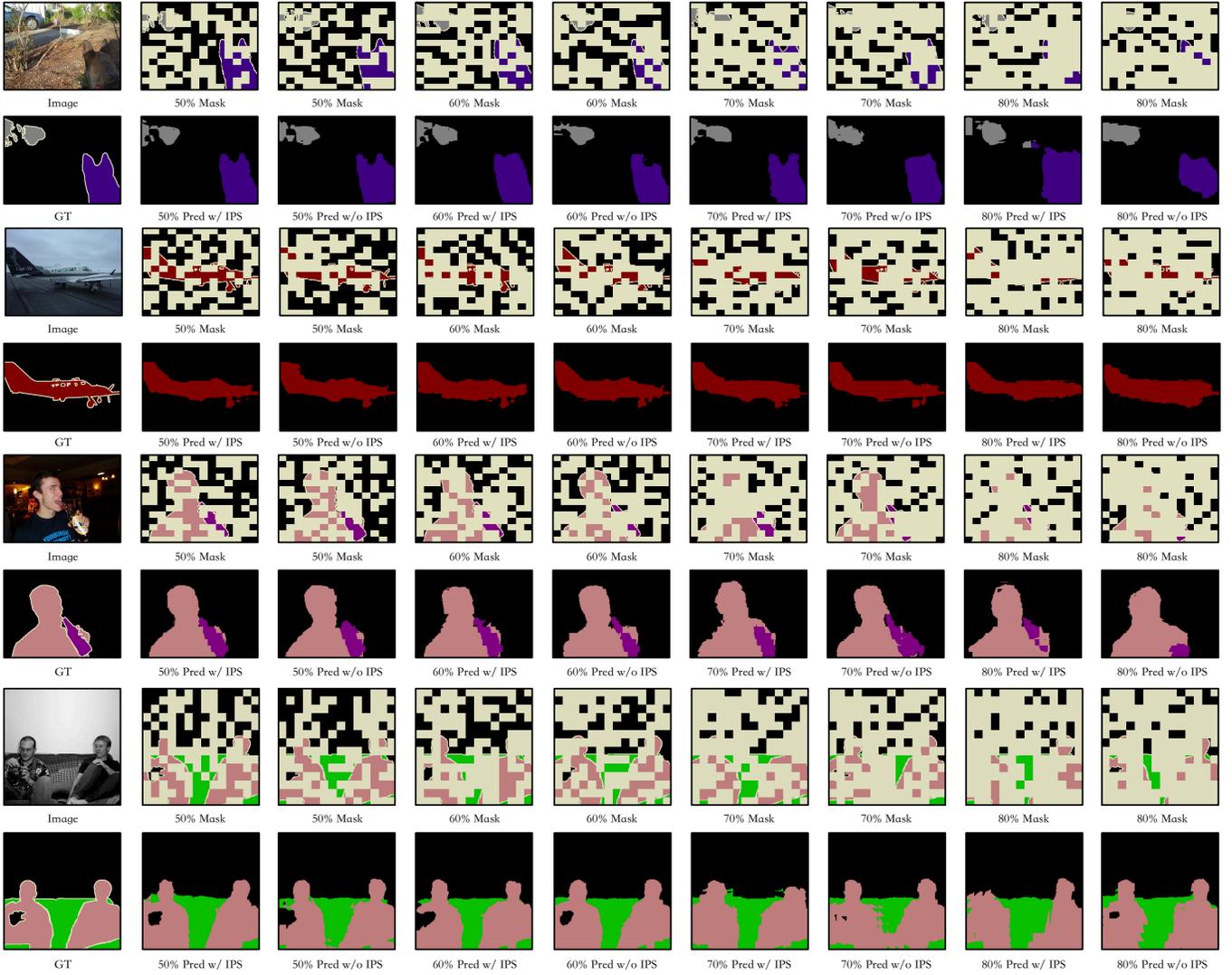}
    \caption{
      Qulitative examples with different settings.
    }
    \label{fig:output2}
    \end{figure*}
    
  \begin{figure*}[tp]
    \centering
    \includegraphics[scale=0.114]{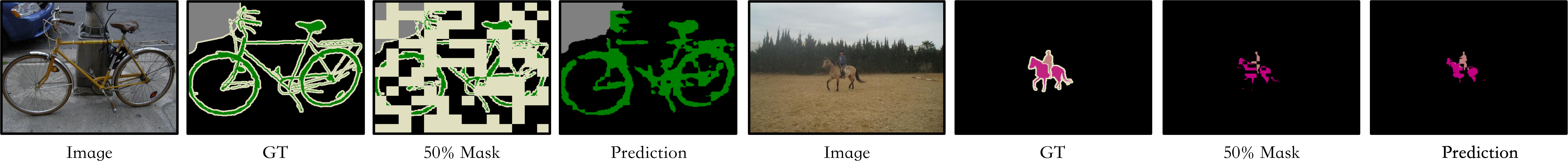}
    \caption{
      Qulitative examples of failure cases.
    }
    \label{fig:outputfail}
    \end{figure*}

 \subsubsection{Mask Ratio \& Image Patch Supplement}
 We have explored the contribution of our Image Patch Supplement algorithm and the effect of Mask Ratio. As shown in \cref{fig:ips}, the mIoU average drop 3.0\%, while the PA-mIoU average drop 4.1\% after applying the Image Patch Supplement(IPS) algorithm. Meanwhile, we notice that the effect of IPS is different under various Mask Ratios. The higher the Mask Ratio, the more significant the improvement of PA-mIoU by IPS. Similarly, the use of IPS has different impacts on mIoU and PA-mIoU. For example, when the Mask Ratio=0.7, the decline of mIoU is smaller than that of PA-mIoU. The result demonstrates that the IPS is an essential algorithm to improve the accuracy, and it proves our IPS can supplement the image information to the property position of the model.
 
\subsubsection{Stack Fuse}
As shown in the table, we compared the performance impact of three different image-label fusion methods on the model. Under the premise that the parameters are set to Encoder Block=8, Decoder Block=6, Encoder Embedding Demension=1440, Decoder Embedding Demension=720, and the patch training strategy is randomly discarded, the fusion method that is directly concating the label to the image, which called Directly Concat, has 72.4\% PA-mIoU and 74.8\% mIoU. The fusion method, in which the label is copied into three copies and inserted into the RGB three of the image, which called Insert Concat, has 75.6\% PA-mIoU and 77.9\% mIoU. In comparison, the method used in this article to layer labels by category, which called Layer Concat, can achieve 94.6\% PA-mIoU and 91.3\% mIoU, fully proving the Advantages of fusion strategies.

 \subsection{Qulitative Study}
 \textbf{Visualization.} As illustrated in \cref{fig:output2}, we present some visualization results with different setting, which demonstrates the benefits of each component in our proposed method. Firstly, under various Mask Ratio setting, compared with the L-MAE employed Image Patch Supplement, the model which without employing perform worse, because the mask process not only discard the label but the image, so the follow-up reconstruct can not use the image information at the masked position. Secondly, as the Mask Ratio raising, the model performance is not affected to much, which proves high stability of the model. Finally, our model can generate high-quality segmentation masks, which demonstrates the effectiveness of our proposed method, i.e., L-MAE.
 
 
 \textbf{Failure Cases.} We visualize some insightful failed cases in \cref{fig:outputfail}. One type of failure is caused when predicting objects containing tubular structures. For the left example in \cref{fig:outputfail}, “green” is not enough to describe the region of the whole bicycle. Besides, for the right example, failures are also caused by the confusion of background information and masked areas in images. At the same time, we also found that when L-MAE completes some small target objects, the completion accuracy will be low. Such a method may need to be solved by reducing the grid size, but it will also increase the number of parameters of the model.
 
 \section{Conclusion}
 
 
 In this paper, we have investigated to leverage the power of Mask AutoEncoder (MAE) models to achieve pixel-level label complete. And, we have proposed an end-to-end Label Mask AutoEncoder (L-MAE) framework to well transfer the mask-reconstruct ability of the MAE model. Compared with the conventional methods, our proposed framework inherits the strong pixel rebuilding ability of the MAE, and the propose method can be used to reconstruct unknown pixels with known label information. The design Image Patch Supplement algorithm can supplement the image information at the desired location, and guarantees the integrity of information when performing completion tasks. In addition, the fusion training algorithm proposed in the article can cover most completion scenarios, making L-MAE better able to complete missing marks in various situations. We conduct comparative experiments on two commonly used datasets, along with extensive ablation studies to validate the effectiveness of each proposed component, and our approach significantly outperforms conventional methods without any pre-training.

  \bibliography{L-MAE}
  \bibliographystyle{IEEEtran}

\end{document}